\documentclass[11pt]{article}

\usepackage[preprint]{acl}

\usepackage{times}
\usepackage{latexsym}

\usepackage[T1]{fontenc}
\usepackage[utf8]{inputenc}

\usepackage{microtype}

\usepackage{inconsolata}

\usepackage{graphicx}
\usepackage{amsmath}
\usepackage{tcolorbox}

\title{Automated Creativity Evaluation for Large Language Models: A Reference-Based Approach}

\author{
  \textbf{Ruizhe Li\textsuperscript{1}}, 
  \textbf{Chiwei Zhu\textsuperscript{1}}, 
  \textbf{Benfeng Xu\textsuperscript{1}}, 
  \textbf{Xiaorui Wang\textsuperscript{2}}, 
  \textbf{Zhendong Mao\textsuperscript{1}} 
  \\
  \textsuperscript{1}University of Science and Technology of China, Hefei, China \\
  \textsuperscript{2}MetastoneTechnology, Beijing, China \\
  \href{mailto:imlrz@mail.ustc.edu.cn}{imlrz@mail.ustc.edu.cn},
  \href{mailto:tanz@mail.ustc.edu.cn}{tanz@mail.ustc.edu.cn}
}

\begin{document}
\maketitle
\begin{abstract}
Creative writing is a key capability of Large Language Models (LLMs), with potential applications in literature, storytelling, and various creative domains. However, evaluating the creativity of machine-generated texts remains a significant challenge, as existing methods either rely on costly manual annotations or fail to align closely with human assessments. In this paper, we propose an effective automated evaluation method based on the Torrance Test of Creative Writing (TTCW), which evaluates creativity as product. Our method employs a reference-based Likert-style approach, scoring generated creative texts relative to high-quality reference texts across various tests. Experimental results demonstrate that our method significantly improves the alignment between LLM evaluations and human assessments, achieving a pairwise accuracy of 0.75 (+15\%).
\end{abstract}

\section{Introduction}

Creative writing is a key capability of Large Language Models (LLMs), with applications in literature, storytelling, and other creative domains \citep{https://doi.org/10.1002/jocb.636, xie-etal-2023-next}. However, studies have revealed a significant gap between the creative writing capabilities of LLMs and those of human experts \citep{ismayilzada2024evaluatingcreativeshortstory, chakrabarty2024art}. Bridging this gap requires further exploration and innovation, which in turn necessitates an effective and practical approach to evaluating the creativity of language models.

Although some studies \citep{stevenson2022puttinggpt3screativityalternative, SummersStay2023BrainstormTS, GUZIK2023100065} have adapted creativity evaluation methods from traditional educational and psychological research—such as the Alternate Uses Task (AUT) \citep{guilford1967creativity} and the Torrance Test of Creative Thinking (TTCT) \citep{torrance1966torrance}—to assess LLMs, these approaches rely heavily on manual annotations. Furthermore, these methods typically evaluate creativity as a process by analyzing responses to open-ended questions designed to elicit creative thinking \citep{cramond2020choosing}, which are inherently difficult to assess automatically. Additionally, the limited number of predefined test questions introduces randomness and increases the likelihood of accidental outcomes\citep{zhao2024assessingunderstandingcreativitylarge}, potentially resulting in unreliable evaluations of LLM performance.

To address these challenges, evaluating creativity as a product rather than a process offers a promising alternative. For instance, \citet{chakrabarty2024art} introduced the Torrance Test of Creative Writing (TTCW), which assesses creativity based on candidates' textual outputs. This approach enhances scalability by allowing the number of test cases to increase continuously while adding the generated texts, thereby reducing randomness through averaging over larger samples. Moreover, automated evaluation of generated texts is more practical compared to subjective judgments of open-ended tasks. However, when applied with LLMs as evaluators, TTCW has not achieved satisfactory results, as reported by \citet{chakrabarty2024art}.

In this paper, we aim to develop an effective automated evaluation method for assessing the creativity of LLMs using TTCW. We draw inspiration from reference-based evaluation methods commonly used in human assessments and automatic evaluations in other fields \citep{zhang2020bertscoreevaluatingtextgeneration, yuan2021bartscoreevaluatinggeneratedtext}, and propose an approach which
assign a relative score to the generated texts compared to high-quality reference texts. Additionally, we adopt Likert-style scoring, a widely used method in psychological assessments, to rate subjective qualities like creativity \citep{roy2020comprehensive}. Experimental results show that our method significantly improves the alignment between LLM evaluations and human assessments, achieving a pairwise accuracy of 0.75 (+15\%).

\section{Related Work}
\subsection{Creativity Evaluation}

In prior work, divergent thinking is widely recognized as a fundamental indicator of creativity in both research and educational settings \citep{Baer1993CreativityAD}. It is typically assessed through open-ended tasks that prompt individuals to generate creative responses. Most widely used methods for evaluating creativity are based on divergent thinking. For example, the Alternate Uses Task (AUT) \citep{guilford1967creativity} asks participants to generate as many novel and unconventional uses as possible for a common object (e.g., a box) within a constrained time period. Similarly, the Torrance Test of Creative Thinking (TTCT)\citep{torrance1966torrance} assesses creativity through responses to novel and unusual scenarios, relying on divergent thinking principles. Our research follows this tradition by grounding creativity evaluation in divergent thinking. Specifically, we adopt the Torrance Test of Creative Writing (TTCW) \citep{chakrabarty2024art}, a variant of TTCT, to evaluate the creativity of LLM-generated texts.

\subsection{Evaluating creativity of large language models}

In recent years, efforts have been made to evaluate the creativity of LLMs. \citep{stevenson2022puttinggpt3screativityalternative} and \citep{GUZIK2023100065} directly apply the Alternate Uses Task (AUT) and the Torrance Test of Creative Thinking (TTCT), respectively. However, both approaches rely heavily on manual annotations, which limit scalability and consistency. Other studies have investigated automated evaluation methods. For example, \citep{beaty2021automating} demonstrated that latent semantic distance is a reliable and strong predictor of human creativity ratings in the AUT. \citep{zhao2024assessingunderstandingcreativitylarge} utilizes GPT-4 to generate TTCT-inspired datasets and employs the model itself to evaluate responses. \citep{chakrabarty2024art} proposes the Torrance Test of Creative Writing (TTCW) and applies it with LLMs as judges though did not yield satisfactory outcomes.

\section{Methodology}

\begin{table*}[ht]
\centering
\begin{tabular}{llccc}
\hline
\textbf{Method} & \textbf{Model} & \textbf{AVG Spearman} & \textbf{AVG Kendall's Tau} & \textbf{Pairwise Accuracy} \\
\hline
Baseline & claudev13        & 0.15  & 0.16  & \textbf{0.64} \\
         & claudev2         & -0.35 & -0.34 & 0.33 \\
         & claudev21        & -0.34 & -0.33 & 0.42 \\
         & claude3-opus     & \textbf{0.25}  & \textbf{0.22}  & \textbf{0.64} \\
         & claude35         & 0.14  & 0.13  & \textbf{0.64} \\
         & cgpt             & -0.40 & -0.38 & 0.36 \\
         & gpt4             & -0.04 & -0.04 & 0.42 \\
         & gpt-4o           & 0.16  & 0.14  & \textbf{0.64} \\
         & gemini-pro       & -0.31 & -0.30 & 0.33 \\
         & qwen2-72b-chat   & -0.12 & -0.11 & 0.58 \\
\hline
Ours     & claude35(ours)   & \textbf{0.49(+0.35)} & \textbf{0.44(+0.31)} & \textbf{0.75(+0.11)} \\
         & gpt-4o(ours)     & 0.38(+0.22)  & 0.36(+0.22)  & 0.72(+0.08) \\
         & qwen2-72b-chat(ours) & 0.22(+0.34)  & 0.16(+0.27)  & 0.61(+0.03) \\
\hline
\end{tabular}
\caption{Comparison of Baseline and Proposed Methods Across Different Models. The table presents the performance of baseline and proposed methods on three metrics: AVG Spearman, AVG Kendall’s Tau, and Pairwise Accuracy. The bolded values in the "Baseline" section represent the highest scores among baseline models. The "Ours" section highlights significant improvements achieved by the proposed method, with changes relative to the baseline shown in parentheses.}
\label{tab:experiment_results}
\end{table*}

\subsection{Problem Setting}

The task of evaluating the creativity of language models is defined as assessing the quality of their generated texts in response to specific prompts. Specifically, plots extracted from human-authored reference stories are used as prompts for the models to generate corresponding stories. The dataset used in this study adopts stories from The New Yorker as the references \citep{chakrabarty2024art}. The process can be denoted as:
$$\text{plot}_i = \text{LLM}_{\text{extract}}(\text{reference}_i)$$
$$\text{candidate}_i^k = \text{LLM}_k(\text{plot}_i)$$
where the reference is a high-quality human-authored story, and $\text{LLM}_k$ represents the model being evaluated.

\subsection{Reference-based Evaluation}

In this evaluation framework, we adopt the Torrance Test of Creative Writing (TTCW) \citep{chakrabarty2024art}, which includes 14 binary tests designed to assess creativity across four dimensions: Fluency, Flexibility, Originality, and Elaboration (see ~\ref{tab:ttcw_dimensions} for details). For each test, the LLM compares the candidate text against the reference text using a Likert scale with five levels: "significantly better" (+2), "slightly better" (+1), "the same" (0), "slightly worse" (-1), and "significantly worse" (-2). To minimize positional bias, the sequence of the candidate and reference texts is alternated, and each test is conducted twice. A test is considered passed (i.e., the test is labeled as "True") if the average score across two assessments is higher than the cutoff score. The overall creativity score of a candidate text is calculated as the total number of tests passed out of the 14 binary tests.

The process is formally represented as:
$$\text{L}_{i,j}^{k,+} = \text{LLM}_{\text{evaluator}}(\text{test}_j, \text{reference}_i, \text{candidate}_i^k)$$
$$\text{L}_{i,j}^{k,-} = \text{LLM}_{\text{evaluator}}(\text{test}_j, \text{candidate}_i^k, \text{reference}_i)$$
$$\text{Score}_i^k = \sum_j I[(\text{L}_{i,j}^{k,+} - \text{L}_{i,j}^{k,-}) \geq \text{score}_{cutoff}]$$
where $\text{L}_{i,k}^{k,+}$ is the label reflecting the extent to which the $\text{candidate}_i^k$ is better than the $\text{reference}_i$, and $\text{L}_{i,k}^{k,-}$ represents the opposite.The $score_{cutoff}$ is a threshold used to convert Likert-scale scores into binary labels, determining whether a candidate passes a given test. A detailed discussion on how the $score_{cutoff}$ is determined and optimized can be found in Discussion Section~\ref{dis_cutoff}.

\subsection{Prompt Strategy}

Previous research has demonstrated that the analyze-rate strategy can improve performance in evaluation tasks when applied with GPT models \citep{chiang2023closerlookautomaticevaluation}. This strategy is similar to zero-shot Chain-of-Thought (CoT) reasoning, but specifically adapted for evaluation tasks. Instead of directly assigning a rating, the model is first prompted to analyze the sample according to the evaluation criteria before providing a final score. In our experiments with different models, We observe the same improvement. Therefore, we adopt this strategy in our final prompt framework, which is detailed in Appendix~\ref{prompt}.

\section{Experiment}

\subsection{Dataset}

This study utilizes the dataset provided by \citep{chakrabarty2024art}, which includes human annotations assessing the creative quality of 12 original stories from The New Yorker alongside corresponding LLM-generated stories produced by GPT-3.5, GPT-4, and Claude V1.3. The statistical details of the dataset can be found in Appendix~\ref{statistics}.

\subsection{Baselines}

For baseline comparisons, we adopt the original prompting method introduced by \citep{chakrabarty2024art} with ten models: Claude 3.5, Claude 3-Opus, Claude V1.3, Qwen-2-72B-Chat, Claude V2.1, Claude V2, GPT-4, GPT-4o, Gemini-Pro, and ChatGPT.

\subsection{Main Result}

In our experiments, we evaluate the effectiveness of the proposed method by its ability to correctly assess the relative capabilities of different models. Specifically, for stories generated from the same plot, we calculate their total scores and derive rankings, which are then compared to rankings provided by human expert evaluators. The ranking similarity is quantified using three metrics: Spearman’s correlation\citep{spearman1904general}, Kendall’s tau\citep{kendall1938new}, and pairwise accuracy, calculated as the proportion of correctly aligned pairwise comparisons between model rankings and human rankings.

As shown in Table~\ref{tab:experiment_results}, our method significantly improves performance among Qwen-2-72B-Chat, GPT-4o, and Claude 3.5. A detailed breakdown of performance across individual stories and models is provided in Appendix ~\ref{result}. Notably, our method achieves the highest pairwise accuracy of 0.75 (+15\%), setting a new benchmark for evaluation reliability.

\section{Discussion}
\subsection{Binary Score Conversion and the Cutoff Score}
\label{dis_cutoff}

To align with the TTCW task setting, which is inherently a binary test, we transformed the raw scores in each tests into binary values. Adopting a binary scoring approach ensures consistency with this format and provides clearer interpretability in downstream tasks. Empirical results further support this decision, showing that ranking similarity improves when using binary scores instead of raw numerical values. 

To determine the optimal cutoff score, we conducted a hyperparameter search, as detailed in Appendix ~\ref{cutoff}. The results indicate that setting the cutoff at -2 yields the best ranking similarity, which aligns with our expectations. A cutoff of -2 corresponds to cases where the average performance across two trials is slightly worse than the reference. Given that reference texts generally exceed the minimum passing standard by a significant margin, we consider candidates who perform only slightly worse than the reference to have still met the test's criteria.

\subsection{Likert Scale Granularity}

To further investigate the impact of Likert scale granularity on experimental results, we conducted an additional study using qwen2-72b-chat to explore different rating scales. Specifically, we evaluated the performance of 3-point, 5-point, and 7-point Likert scales to determine the optimal level of granularity for our evaluation framework. The Spearman’s correlation for the 5-point scale is 0.22, outperforming both the 3-point scale (-0.07) and the 7-point scale (0.01). These findings suggest that the 5-point Likert scale is a more effective choice for our evaluation framework.

\subsection{Ablation Study}

To evaluate the impact of the Reference-Based Approach and the Analyze-Rate Strategy on the evaluation framework, we conducted ablation experiments by separately removing each component. In the ablation of the Reference-Based Approach, we removed the reference-based comparison, instructing the LLM to assess the candidate text solely based on its content and generate a binary label at the end of its response. In the ablation of the Analyze-Rate Strategy, we removed the analyze-rate prompting method and prompted the LLM to assign a label directly, without an explicit instruction to analyze the sample before rating.

The results, detailed in the ~\ref{ablation}, indicate that both the Reference-Based Approach and the Analyze-Rate Strategy significantly enhance evaluation performance. Removing either component led to a decrease in ranking similarity and evaluation stability, highlighting the complementary roles of both strategies in ensuring a robust and reliable automated evaluation system.

\subsection{Robustness and Generalization Ability}

To further assess the robustness and generalization of our evaluation framework, we conducted experiments on the dataset introduced by \citep{gómezrodríguez2023confederacymodelscomprehensiveevaluation}, which presents a different evaluation setup compared to TTCW. While TTCW employs a binary label for each test, this dataset utilizes a 10-point rating scale. Despite this fundamental difference, our framework—still operating with binary labels—exhibited strong alignment with human ratings, as provided in Appendix~\ref{result2}, demonstrating its ability to generalize effectively beyond binary classification settings.

Another notable distinction in this dataset is the absence of human expert-authored texts, as the highest-scoring texts are generated by GPT-4 rather than human writers. Consequently, in our evaluation, GPT-4-generated texts were used as references. The dataset itself comprises five distinct storylines, each written by ten different language models or humans, resulting in a total of 50 generated texts. The strong performance of our framework across this dataset further underscores its applicability beyond TTCW, suggesting its potential for broader evaluation tasks. For instance, this approach could be extended to assessing the creativity of distilled models relative to their teacher models, as well as other tasks requiring automated model evaluation.

\section{Conclusion}

We proposed an automated evaluation framework for assessing the creativity of large language models (LLMs) using the Torrance Test of Creative Writing (TTCW). By adopting a reference-based Likert-style evaluation and an analyze-rate prompting strategy, our method improves alignment with human assessments, achieving a pairwise accuracy of 0.75 (+15\%). Ablation studies highlight the complementary roles of the reference-based approach and analyze-rate prompting, while experiments on a 10-point scale dataset with GPT-4 references demonstrate its robustness and generalizability. These results establish a new benchmark for automated creativity evaluation, offering a scalable alternative to manual annotation.

\section{Limitation}
One limitation of our method is its reliance on reference stories, which may restrict its scalability for unrestricted article-level evaluations. Additionally, our method may not be suitable when all candidate texts are far inferior to the reference, as this could result in all labels being assigned as significantly worse, making it impossible to distinguish relative rankings among candidates. Nonetheless, this approach serves as a robust framework for comparing the creative capabilities of different models, providing valuable insights into their relative performance.

\section{Potential Risks}

The proposed evaluation framework, while promising, carries potential risks that may impact its broader application and outcomes. One concern is amplifying biases in reference texts, which could favor certain styles or cultural norms while disadvantaging unconventional outputs. Additionally, automating creativity evaluation risks reducing human oversight, potentially overlooking nuanced, subjective aspects of creativity that machines cannot fully capture. Addressing these challenges requires careful reference selection and maintaining a balance between automated and human evaluations.

\bibliography{main}

\begin{thebibliography}{20}
\providecommand{\natexlab}[1]{#1}

\bibitem[{Baer(1993)}]{Baer1993CreativityAD}
John Baer. 1993.
\newblock \href {https://api.semanticscholar.org/CorpusID:145663312} {Creativity and divergent thinking: A task-specific approach}.

\bibitem[{Beaty and Johnson(2021)}]{beaty2021automating}
Roger~E. Beaty and Dan~R. Johnson. 2021.
\newblock Automating creativity assessment with semdis: An open platform for computing semantic distance.
\newblock \emph{Behavior Research Methods}, 53(2):757--780.

\bibitem[{Chakrabarty et~al.(2024)Chakrabarty, Laban, Agarwal, Muresan, and Wu}]{chakrabarty2024art}
Tuhin Chakrabarty, Philippe Laban, Divyansh Agarwal, Smaranda Muresan, and Chien-Sheng Wu. 2024.
\newblock Art or artifice? large language models and the false promise of creativity.
\newblock In \emph{Proceedings of the CHI Conference on Human Factors in Computing Systems}, pages 1--34.

\bibitem[{Chiang and yi~Lee(2023)}]{chiang2023closerlookautomaticevaluation}
Cheng-Han Chiang and Hung yi~Lee. 2023.
\newblock \href {https://arxiv.org/abs/2310.05657} {A closer look into automatic evaluation using large language models}.
\newblock \emph{Preprint}, arXiv:2310.05657.

\bibitem[{Cramond(2020)}]{cramond2020choosing}
B~Cramond. 2020.
\newblock Choosing a creativity assessment that is fit for purpose.
\newblock \emph{Assessing Creativity: A palette of possibilities}, pages 58--63.

\bibitem[{Guilford(1967)}]{guilford1967creativity}
J.P. Guilford. 1967.
\newblock Creativity: Yesterday, today and tomorrow.
\newblock \emph{The Journal of Creative Behavior}, 1(1):3--14.

\bibitem[{Guzik et~al.(2023)Guzik, Byrge, and Gilde}]{GUZIK2023100065}
Erik~E. Guzik, Christian Byrge, and Christian Gilde. 2023.
\newblock \href {https://doi.org/10.1016/j.yjoc.2023.100065} {The originality of machines: Ai takes the torrance test}.
\newblock \emph{Journal of Creativity}, 33(3):100065.

\bibitem[{Gómez-Rodríguez and Williams(2023)}]{gómezrodríguez2023confederacymodelscomprehensiveevaluation}
Carlos Gómez-Rodríguez and Paul Williams. 2023.
\newblock \href {https://arxiv.org/abs/2310.08433} {A confederacy of models: a comprehensive evaluation of llms on creative writing}.
\newblock \emph{Preprint}, arXiv:2310.08433.

\bibitem[{Ismayilzada et~al.(2024)Ismayilzada, Stevenson, and van~der Plas}]{ismayilzada2024evaluatingcreativeshortstory}
Mete Ismayilzada, Claire Stevenson, and Lonneke van~der Plas. 2024.
\newblock \href {https://arxiv.org/abs/2411.02316} {Evaluating creative short story generation in humans and large language models}.
\newblock \emph{Preprint}, arXiv:2411.02316.

\bibitem[{Kendall(1938)}]{kendall1938new}
Maurice~G. Kendall. 1938.
\newblock A new measure of rank correlation.
\newblock \emph{Biometrika}, 30(1-2):81--93.

\bibitem[{Orwig et~al.(2024)Orwig, Edenbaum, Greene, and Schacter}]{https://doi.org/10.1002/jocb.636}
William Orwig, Emma~R. Edenbaum, Joshua~D. Greene, and Daniel~L. Schacter. 2024.
\newblock \href {https://doi.org/10.1002/jocb.636} {The language of creativity: Evidence from humans and large language models}.
\newblock \emph{The Journal of Creative Behavior}, 58(1):128--136.

\bibitem[{Roy(2020)}]{roy2020comprehensive}
A.~Roy. 2020.
\newblock \href {https://books.google.de/books?id=sWQOzgEACAAJ} {\emph{A Comprehensive Guide for Design, Collection, Analysis and Presentation of Likert and Other Rating Scale Data: Analysis of Likert Scale Data}}.
\newblock Amazon Digital Services LLC - KDP Print US.

\bibitem[{Spearman(1904)}]{spearman1904general}
Charles Spearman. 1904.
\newblock {``General Intelligence,'' Objectively Determined and Measured}.
\newblock \emph{The American Journal of Psychology}, 15(2):201--292.

\bibitem[{Stevenson et~al.(2022)Stevenson, Smal, Baas, Grasman, and van~der Maas}]{stevenson2022puttinggpt3screativityalternative}
Claire Stevenson, Iris Smal, Matthijs Baas, Raoul Grasman, and Han van~der Maas. 2022.
\newblock \href {https://arxiv.org/abs/2206.08932} {Putting gpt-3's creativity to the (alternative uses) test}.
\newblock \emph{Preprint}, arXiv:2206.08932.

\bibitem[{Summers-Stay et~al.(2023)Summers-Stay, Lukin, and Voss}]{SummersStay2023BrainstormTS}
Douglas Summers-Stay, Stephanie~M. Lukin, and Clare~R. Voss. 2023.
\newblock \href {https://api.semanticscholar.org/CorpusID:259305709} {Brainstorm, then select: a generative language model improves its creativity score}.

\bibitem[{Torrance(1966)}]{torrance1966torrance}
E~Paul Torrance. 1966.
\newblock Torrance tests of creative thinking.
\newblock \emph{Educational and psychological measurement}.

\bibitem[{Xie et~al.(2023)Xie, Cohn, and Lau}]{xie-etal-2023-next}
Zhuohan Xie, Trevor Cohn, and Jey~Han Lau. 2023.
\newblock \href {https://doi.org/10.18653/v1/2023.inlg-main.23} {The next chapter: A study of large language models in storytelling}.
\newblock In \emph{Proceedings of the 16th International Natural Language Generation Conference}, pages 323--351, Prague, Czechia. Association for Computational Linguistics.

\bibitem[{Yuan et~al.(2021)Yuan, Neubig, and Liu}]{yuan2021bartscoreevaluatinggeneratedtext}
Weizhe Yuan, Graham Neubig, and Pengfei Liu. 2021.
\newblock \href {https://arxiv.org/abs/2106.11520} {Bartscore: Evaluating generated text as text generation}.
\newblock \emph{Preprint}, arXiv:2106.11520.

\bibitem[{Zhang et~al.(2020)Zhang, Kishore, Wu, Weinberger, and Artzi}]{zhang2020bertscoreevaluatingtextgeneration}
Tianyi Zhang, Varsha Kishore, Felix Wu, Kilian~Q. Weinberger, and Yoav Artzi. 2020.
\newblock \href {https://arxiv.org/abs/1904.09675} {Bertscore: Evaluating text generation with bert}.
\newblock \emph{Preprint}, arXiv:1904.09675.

\bibitem[{Zhao et~al.(2024)Zhao, Zhang, Li, Huang, Guo, Peng, Hao, Wen, Hu, Du, Guo, Li, and Chen}]{zhao2024assessingunderstandingcreativitylarge}
Yunpu Zhao, Rui Zhang, Wenyi Li, Di~Huang, Jiaming Guo, Shaohui Peng, Yifan Hao, Yuanbo Wen, Xing Hu, Zidong Du, Qi~Guo, Ling Li, and Yunji Chen. 2024.
\newblock \href {https://arxiv.org/abs/2401.12491} {Assessing and understanding creativity in large language models}.
\newblock \emph{Preprint}, arXiv:2401.12491.

\end{thebibliography}

\clearpage

\onecolumn
\appendix

\section{Appendix}

\subsection{Prompt}
\label{prompt}
In this section, we provide the prompt used to generate the evaluation results.
\begin{tcolorbox}
Please act as an experienced and impartial literary critic to evaluate the creativity of two stories. You will be provided with two stories, Story A and Story B. You will then be given some background knowledge on specific aspects of creative writing. Carefully read both stories and, using the provided background knowledge, critically analyze them for their creativity.\\ \\
Think step by step, and describe your thought process using concise phrases. After providing your analysis, you must conclude by outputting only one of the following choices as your final verdict with a label:\\ \\
1. Story A is significantly better: [[A>>B]]\\
2. Story A is slightly better: [[A>B]]\\
3. Tie, relatively the same: [[A=B]]\\
4. Story B is slightly better: [[B>A]]\\
5. Story B is significantly better: [[B>>A]]\\ \\
Example output: "A: narrative ending, ... B: poor character development, ... Therefore: [[A>B]]".\\
\\
Stories and Question...\\
\\
Remember, you must end your answer with one of these: [[A>>B]], [[A>B]], [[A=B]], [[B>A]], [[B>>A]]
\end{tcolorbox}

\clearpage

\subsection{Dataset Statistics and Additional Results}
\label{statistics}
\subsubsection{Word Counts for Different Models}
To provide further insights into the dataset, Table~\ref{tab:word_counts} presents the word counts of generated stories across different models. While differences in verbosity and writing style exist, stories generated from the same storyline tend to have similar word counts, reducing the potential impact of length variations on evaluation scores.

\begin{table}[h]
\centering
\caption{Word counts of generated stories for different models. The New Yorker column represents the original human-written reference texts.}
\label{tab:word_counts}
\begin{tabular}{lrrrr}
\hline
\textbf{Story Name} & \textbf{Claude} & \textbf{GPT-3.5} & \textbf{GPT-4} & \textbf{New Yorker} \\
\hline
A Triangle                              &     831 &    1126 &  1074 &        959  \\
Barbara, Detroit, 1996                   &    1245 &    1452 &  1460 &       1432  \\
Beyond Nature                           &    1245 &    1628 &  1326 &       1476  \\
Certain European Movies                 &    1304 &    1623 &  1480 &       1584  \\
Keys                                    &    1370 &    1630 &  1297 &       1433  \\
Listening For the Click                 &    1463 &    1623 &  1612 &       1467  \\
Maintenance, Hvidovre                   &    1270 &    1992 &  1911 &       2066  \\
Returns                                 &    1519 &    1726 &  1765 &       1715  \\
The Facade Renovation That's Going Well &    1332 &    1544 &  1477 &       1501  \\
The Kingdom That Failed                 &    1344 &    1344 &  1356 &       1525  \\
The Last Dance with my Dad              &    1406 &    2455 &  1932 &       2233  \\
Trash                                   &    1541 &    2215 &  2398 &       2350  \\
\hline
\end{tabular}
\end{table}

\subsubsection{TTCW Score Distribution}
\begin{figure}[h]
    \centering
    \includegraphics[width=0.95\textwidth]{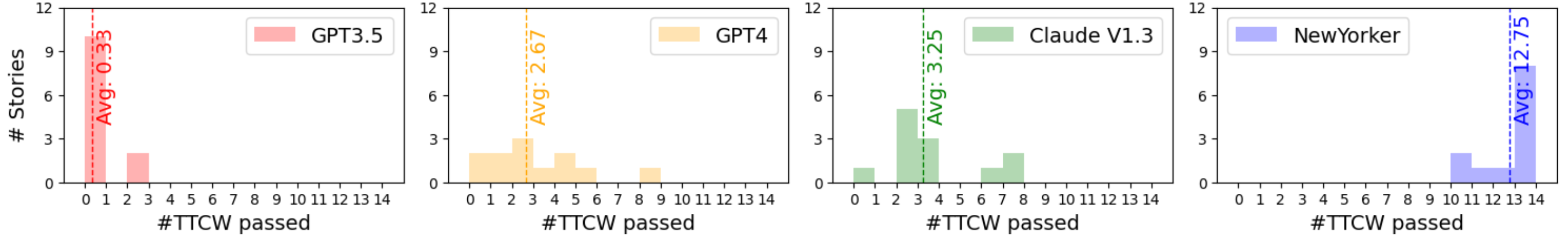}
    \caption{Distribution of TTCW test scores across different models. The dashed lines indicate the average number of tests passed by each model.}
    \label{fig:ttcw_scores}
\end{figure}

Figure~\ref{fig:ttcw_scores} presents the distribution of TTCW test scores across different models. Each histogram represents the number of stories that passed a given number of tests, along with the corresponding average score.

\clearpage

\subsection{TTCW Test}
\label{tab:ttcw_dimensions}
This section presents the TTCW test, which outlines the dimensions and guiding questions for evaluating creativity in stories. The test includes four key dimensions: fluency, flexibility, originality, and elaboration, each accompanied by detailed background knowledge to facilitate a structured analysis. The Torrance Test of Creative Writing (TTCW) is distributed under the BSD-3-Clause license.
\begin{table}[htbp]
\centering
\caption{TTCW Dimensions and Questions}
\label{tab:ttcw_test}
\renewcommand{\arraystretch}{1.3} % Adjust row spacing
\begin{tabular}{|p{2.5cm}|p{10.5cm}|}
\hline
\textbf{Dimension} & \textbf{Question} \\
\hline
Fluency & Does the end of the story feel natural and earned, as opposed to arbitrary or abrupt? \\
\hline
Fluency & Do the different elements of the story work together to form a unified, engaging, and satisfying whole? \\
\hline
Fluency & Does the story have an appropriate balance between scene and summary/exposition, or does it rely too heavily on one element? \\
\hline
Fluency & Does the manipulation of time (compression or stretching) feel appropriate and balanced? \\
\hline
Fluency & Does the story make sophisticated use of idiom, metaphor, or literary allusion? \\
\hline
Flexibility & Does the story achieve a good balance between interiority and exteriority, in a way that feels emotionally flexible? \\
\hline
Flexibility & Does the story contain turns that are both surprising and appropriate? \\
\hline
Flexibility & Does the story provide diverse perspectives, and if there are unlikeable characters, are their perspectives presented convincingly and accurately? \\
\hline
Originality & Is the story an original piece of writing without any clichés? \\
\hline
Originality & Does the story show originality in its form and/or structure? \\
\hline
Originality & Will an average reader of this story obtain a unique and original idea from reading it? \\
\hline
Elaboration & Are there passages in the story that involve subtext, and if so, does the subtext enrich the setting or feel forced? \\
\hline
Elaboration & Does the writer make the fictional world believable at the sensory level? \\
\hline
Elaboration & Does each character feel developed with appropriate complexity, ensuring no character exists solely for plot convenience? \\
\hline
\end{tabular}
\end{table}

\clearpage
\subsection{Full Result}
\label{result}

\begin{figure*}[h!]
  \centering
  \includegraphics[width=\textwidth]{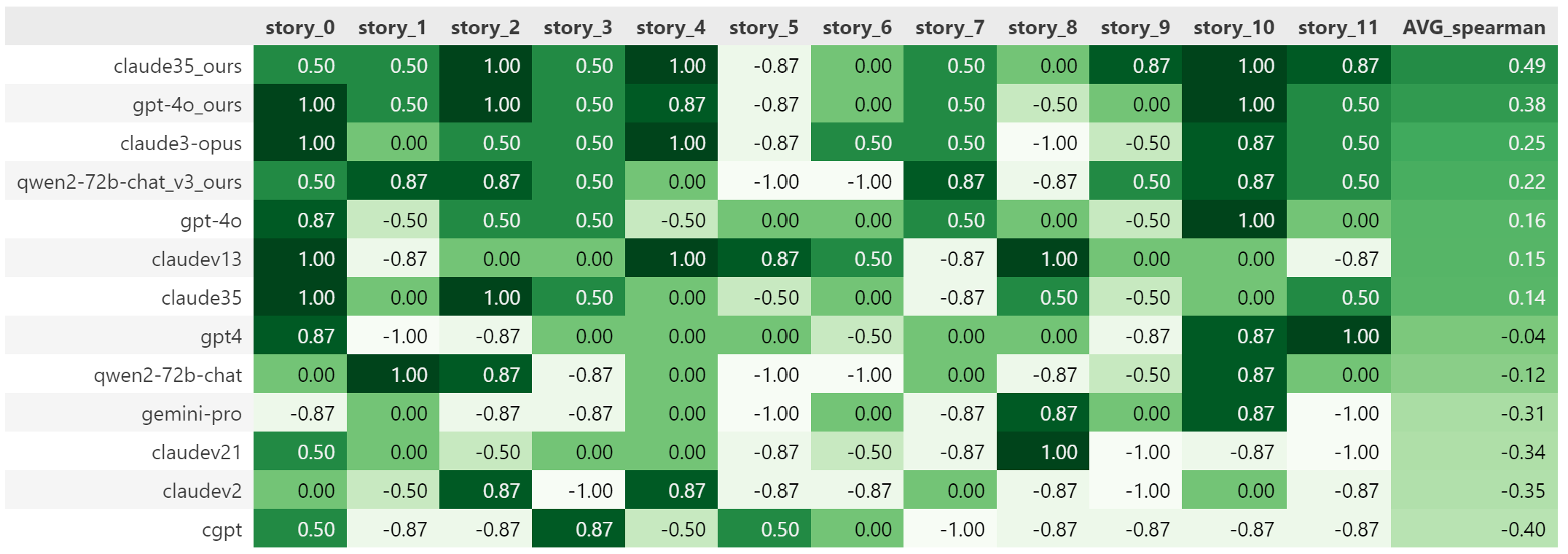}
  \caption{Complete Spearman correlation results across individual stories and models. Models labeled 'ours' indicate performance using our proposed method. The results are sorted in descending order of the average values.}
  \label{fig:sp}
\end{figure*}

\begin{figure*}[h!]
  \centering
  \includegraphics[width=\textwidth]{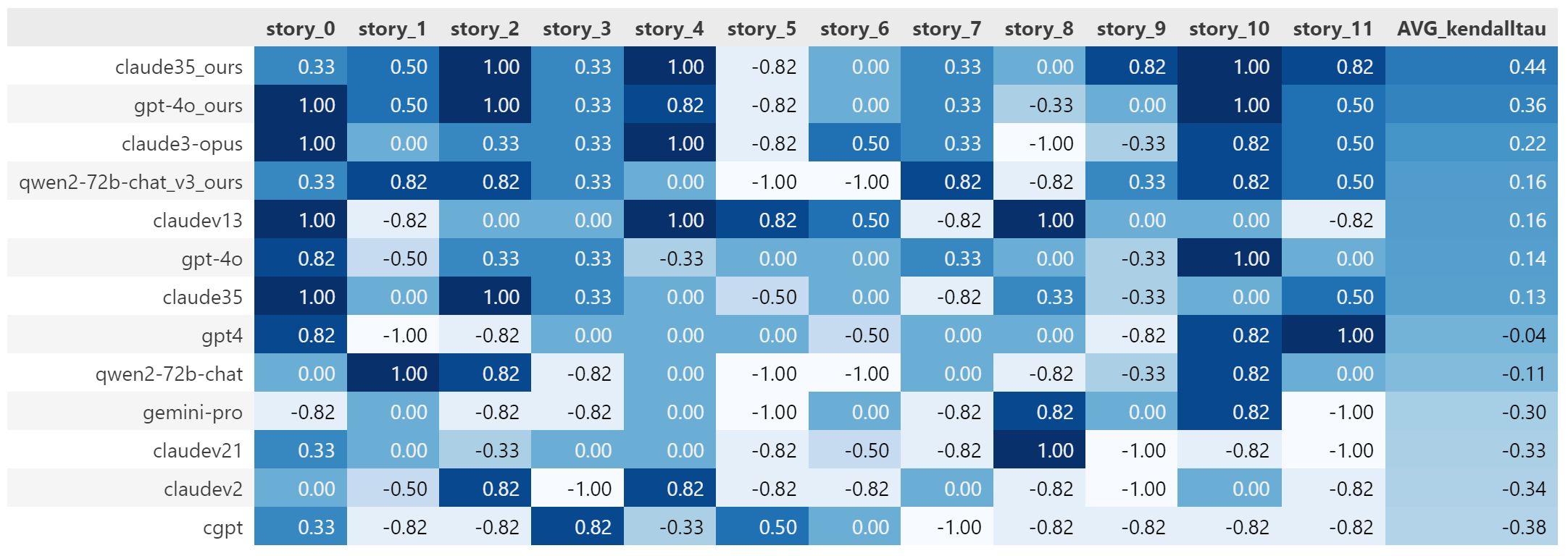}
  \caption{Complete Kendall’s tau results across individual stories and models. Models labeled 'ours' indicate performance using our proposed method. The results are sorted in descending order of the average values.}
  \label{fig:kt}
\end{figure*}

\begin{figure*}[h!]
  \centering
  \includegraphics[width=\textwidth]{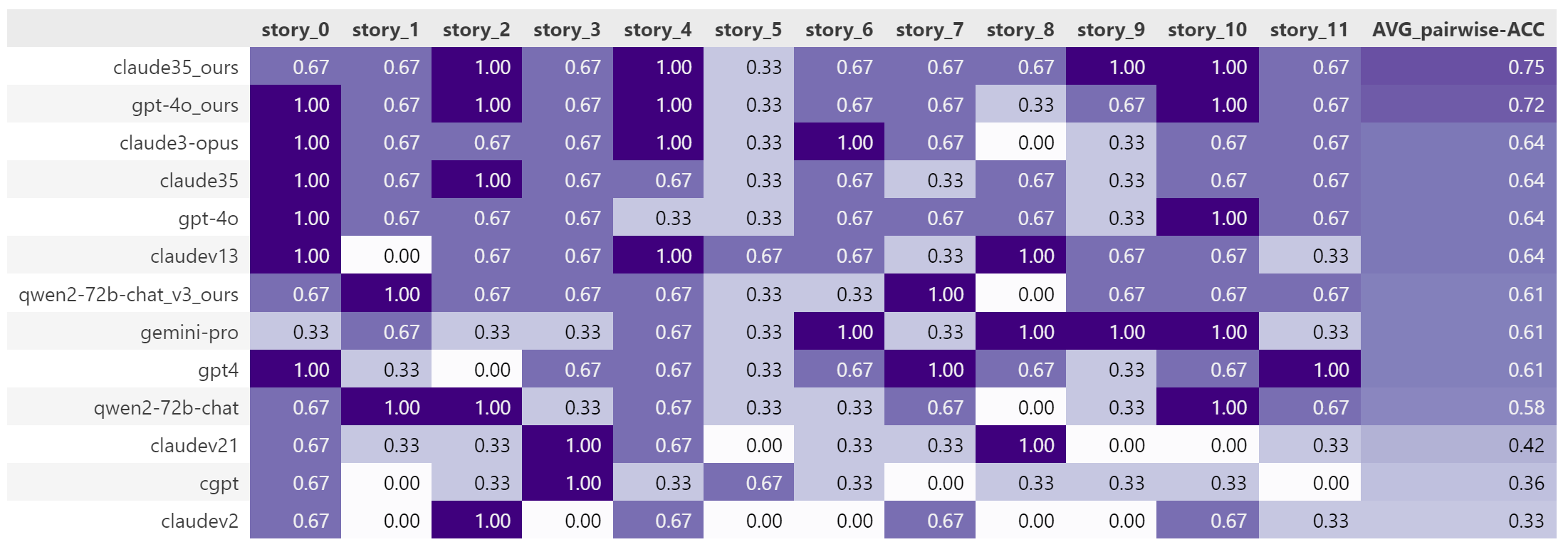}
  \caption{Complete Pairwise accuracy results across individual stories and models. Models labeled 'ours' indicate performance using our proposed method. The results are sorted in descending order of the average values.}
  \label{fig:pa}
\end{figure*}

\clearpage
\subsection{Results Obtained with Different Cutoff Scores}
\label{cutoff}

\begin{table}[h]
\centering
\begin{tabular}{lcccc}
\hline
\textbf{Model} & \textbf{Cutoff = -3} & \textbf{Cutoff = -2} & \textbf{Cutoff = -1} & \textbf{Cutoff = 0} \\
\hline
Qwen            & -0.12  & -0.12  & -0.12  & -0.12  \\
Qwen-Ours       & 0.17   & 0.22   & -0.05  & -0.01  \\
GPT-4o          & 0.16   & 0.16   & 0.16   & 0.16   \\
GPT-4o-Ours     & 0.27   & 0.38   & 0.33   & 0.22   \\
Claude 3.5      & 0.14   & 0.14   & 0.14   & 0.14   \\
Claude 3.5-Ours & 0.20   & 0.49   & 0.37   & 0.30   \\
\hline
\end{tabular}
\caption{Spearman correlation of different models under varying cutoff scores.}
\label{tab:cutoff_experiment}
\end{table}

\begin{figure}[h]
    \centering
    \includegraphics[width=0.75\textwidth]{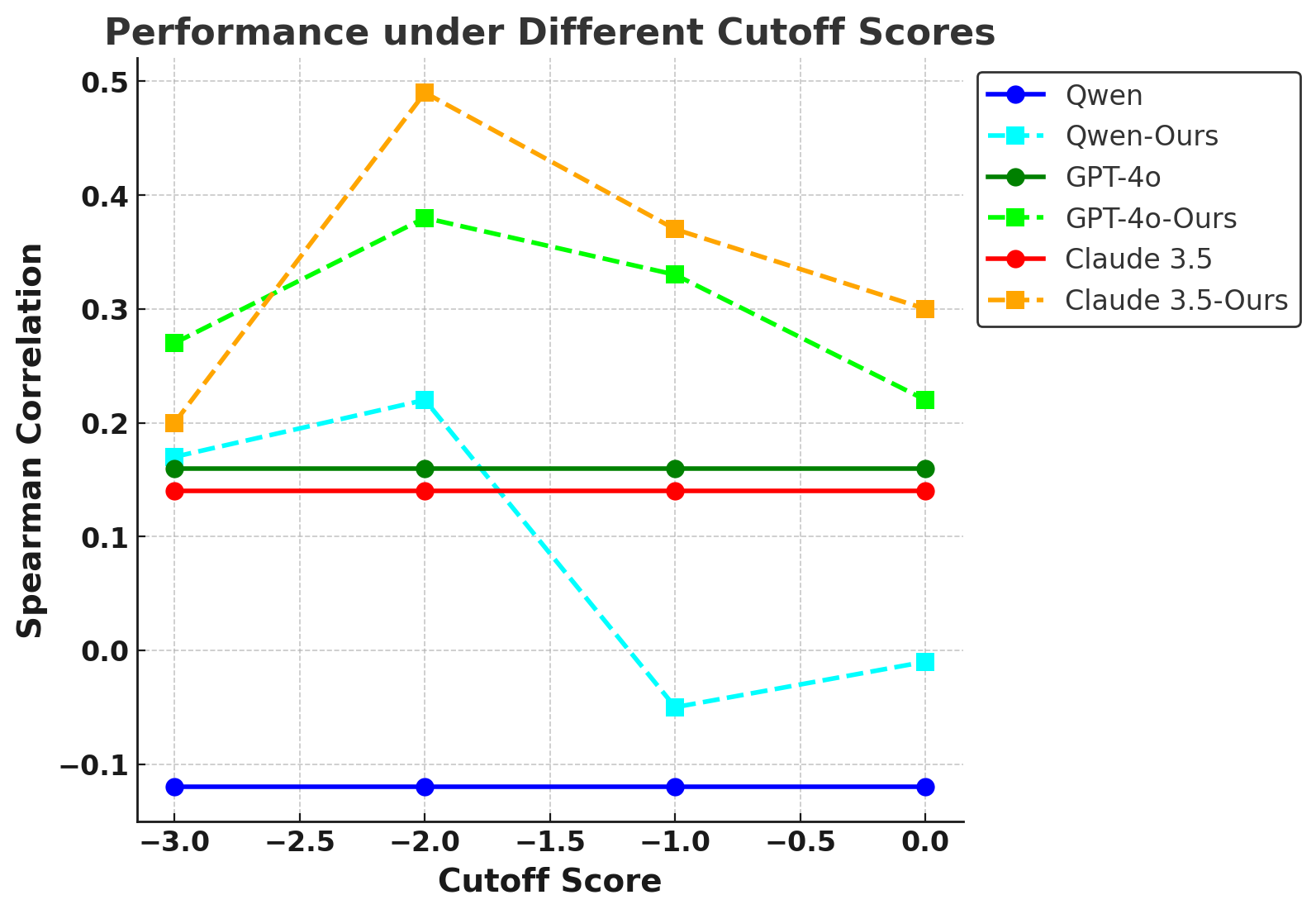}
    \caption{Spearman correlation performance under different cutoff scores.}
    \label{fig:cutoff_plot}
\end{figure}

\subsection{Ablation Study}
\label{ablation}
\begin{table}[h]
\centering
\renewcommand{\arraystretch}{1.2} % 增加行距，提高可读性
\begin{tabular}{lcc}
\hline
\textbf{Method} & \textbf{Qwen2-72B-Chat} & \textbf{Claude 3.5} \\
\hline
Ours                              & \textbf{0.22}  & \textbf{0.49}  \\
Reference-Based Approach Only      & 0.00  & 0.42  \\
Analyze-Rate Prompting Only        & 0.16  & 0.45  \\
Baseline                           & -0.12 & 0.14  \\
\hline
\end{tabular}
\caption{Ablation study results showing Spearman’s correlation ($\rho$) for different evaluation strategies. The best performance for each model is in \textbf{bold}.}
\label{tab:ablation_experiment}
\end{table}

\clearpage
\subsection{Full Result on Additional Dataset}
\label{result2}

\begin{figure*}[h!]
  \centering
  \includegraphics[width=\textwidth]{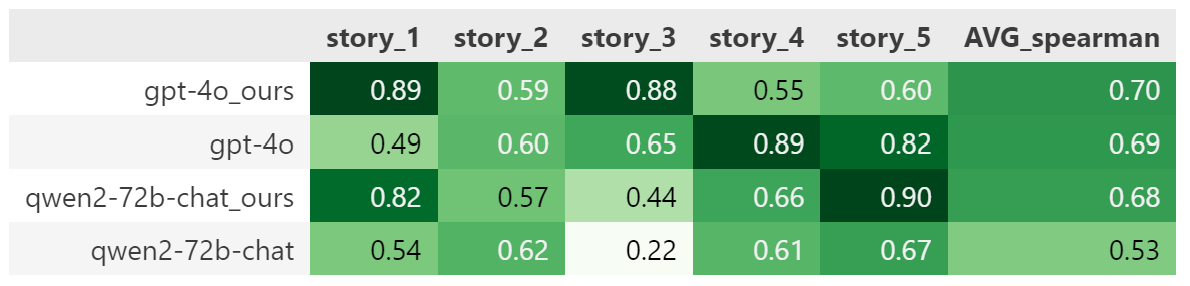}
  \caption{Complete Spearman correlation results across individual stories and models. Models labeled 'ours' indicate performance using our proposed method. The results are sorted in descending order of the average values.}
  \label{fig:sp}
\end{figure*}

\begin{figure*}[h!]
  \centering
  \includegraphics[width=\textwidth]{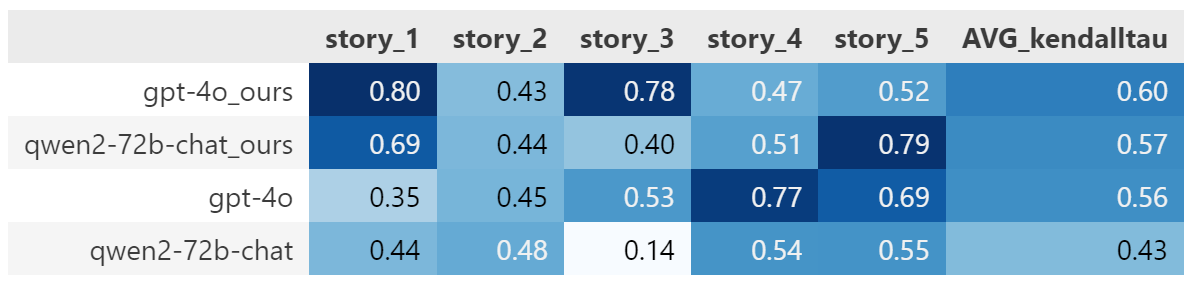}
  \caption{Complete Kendall’s tau results across individual stories and models. Models labeled 'ours' indicate performance using our proposed method. The results are sorted in descending order of the average values.}
  \label{fig:kt}
\end{figure*}

\begin{figure*}[h!]
  \centering
  \includegraphics[width=\textwidth]{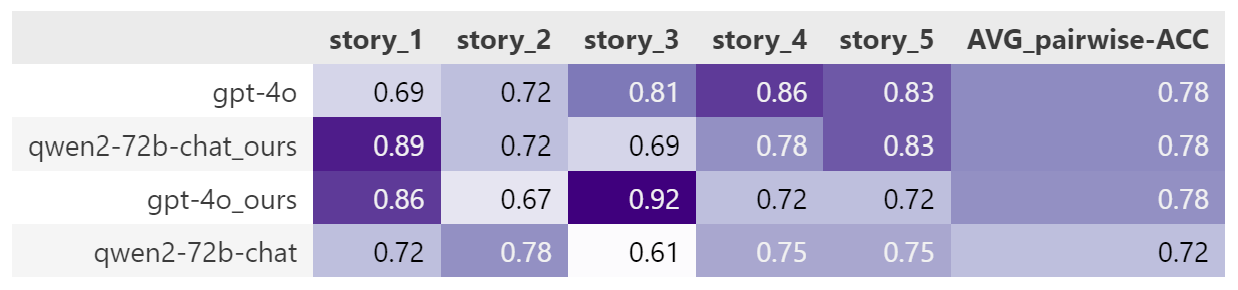}
  \caption{Complete Pairwise accuracy results across individual stories and models. Models labeled 'ours' indicate performance using our proposed method. The results are sorted in descending order of the average values.}
  \label{fig:pa}
\end{figure*}

\end{document}